%% file: main.tex
\documentclass[11pt]{article}

\usepackage[final]{acl}

\usepackage{times}
\usepackage{latexsym}

\usepackage{enumitem}
\usepackage[T1]{fontenc}
\usepackage[utf8]{inputenc}
\usepackage{amsfonts}
\usepackage{amsmath}
\usepackage{xurl}
\usepackage{booktabs}
\usepackage{threeparttable}
\usepackage{siunitx}
\usepackage{multirow}
\usepackage{microtype}

\usepackage{inconsolata}

\usepackage{graphicx}

\title{Before the Warning Comes Too Late: Incremental Phone-Scam Detection from Speech}

\author{
  \textbf{Khang Nhat Hoang Vo}\textsuperscript{1,2}\thanks{Work done while visiting the National University of Singapore.} \quad
  \textbf{Anh Trac Duc Dinh}\textsuperscript{3,4} \quad
  \textbf{Tai Tien Ta\textsuperscript{4}} \quad
  \textbf{Tho Quan\textsuperscript{4}}
\\
\\
  \textsuperscript{1}Mohamed bin Zayed University of Artificial Intelligence, Abu Dhabi, UAE
  \\
  \textsuperscript{2}National University of Singapore, Singapore
  \\
   \textsuperscript{3}Center for AI Reseach (CAIR), VinUniversity, Hanoi
  \\ 
  \textsuperscript{4}Faculty of Computer Science and Engineering, Ho Chi Minh City University of Technology \\(HCMUT), VNU-HCM, Ho Chi Minh City, Vietnam
  \\
  \small{\textbf{Correspondence:} \href{mailto:Khang.Vo@mbzuai.ac.ae}{Khang.Vo@mbzuai.ac.ae}, \href{mailto:qttho@hcmut.edu.vn}{qttho@hcmut.edu.vn}
  }
}

\begin{document}
\maketitle

\input{sections/abstract}
\input{sections/introduction}
\input{sections/related_work}
\input{sections/methodology}
\input{sections/experiment_result}

\input{sections/conclusion}

\section*{Limitations}

The controlled English benchmark is constructed from scripted dialogues and
synthetic speech. Although this design supports balanced and reproducible
evaluation, it does not capture the full variability of real calls, including
spontaneous speech, interruptions, background noise, channel distortion,
regional accents, and evolving scam strategies. Evaluation on
TeleAntiFraud-28K extends the study to Mandarin conversations, but the
comparison with published Qwen2-Audio systems is contextual rather than
controlled because the models use different inputs, supervision, and
training procedures. StreamFraudNet is trained only with conversation-level labels. Consequently,
its window-level outputs are latent risk scores rather than validated
localizations of fraudulent evidence. The prefix experiments show that useful
predictions are available before a call ends, but they do not measure delay
relative to the true onset of fraudulent content. Moreover, the matched
global BiLSTM remains numerically stronger at every evaluated prefix.
Timestamped annotations and latency-aware objectives are therefore needed to
evaluate and improve fraud-onset detection directly.

The efficiency results are limited to the evaluated server GPU and CPU.
They establish faster-than-real-time processing on these platforms, but not
deployment on mobile phones, embedded processors, or telecommunication edge
devices. Although the task-specific head contains only 0.953 million
trainable parameters, the complete inference pipeline contains 95.325
million parameters because the frozen speech encoder remains required.

Finally, raw telephone audio may contain sensitive personal information.
Practical deployment would require appropriate consent, secure processing
and storage, restricted retention, and validation across languages,
demographic groups, and acoustic conditions. Fraud predictions should
support human or policy-based review rather than automatically blocking
calls, since false positives may disrupt legitimate communication and false
negatives may leave users unprotected.

\bibliography{ref}

\include{sections/appendix}

\end{document}

%% file: sections/abstract.tex
\begin{abstract}
We study weakly supervised incremental telecom fraud detection from raw
telephone audio, where training provides only conversation-level labels and
predictions must be updated before a call ends. We introduce StreamFraudNet,
which processes incoming audio through overlapping bounded-context windows
using a frozen self-supervised speech encoder, recurrent temporal modeling,
and learned aggregation of latent window scores. On a controlled English
benchmark, StreamFraudNet achieves a ROC--AUC of \(0.9953\), significantly
outperforming acoustic and mean-pooling baselines while remaining competitive
with strong global temporal models. The model produces its first prediction
after 10 seconds of audio, updates every 2 seconds, and operates faster than
real time on the evaluated server hardware. Ablations identify recurrent
temporal context as the principal contributor to performance. These results
demonstrate that fraud risk can be scored incrementally from raw speech
without transcripts or temporal annotations, while highlighting the need for
latency-aware training to improve early prediction.
\end{abstract}

%% file: sections/introduction.tex
\section{Introduction}
\label{sec:intro}

Phone-scam detection is inherently time-sensitive. Recent LLM-based systems
analyze ongoing calls and issue warnings before harmful actions are completed
~\citep{shen2025where,shen2025rightmoment,10.1145/3715070.3757231,10902894}.
This motivates a setting beyond post-call classification: estimating fraud
risk from a partially observed conversation and updating that estimate as new
evidence arrives. The need is substantial. Global telecom fraud losses reached
USD~\(38.95\) billion in 2023~\citep{cfca2023fraudloss}, while recent work
shows how language models, speech recognition, and speech synthesis can be
combined into scalable scam pipelines
~\citep{gressel2024scamautomation,unodc2025inflection}.

Existing conversational fraud detectors commonly rely on ASR transcripts,
large audio--language models, or richer reasoning-oriented supervision
~\citep{wang2026safeqaq,10.1145/3719027.3765567,
chen-etal-2026-detecting,shen2025where,shen2025rightmoment,
ma2025teleantifraud}. These approaches can model semantic content effectively,
but may require an intermediate transcription stage, high-capacity models, or
fine-grained annotations. We instead study whether fraud risk can be updated
directly from raw telephone audio using only binary conversation-level labels.

This setting is weakly supervised because the model is told whether a call is
fraudulent, but not when fraudulent evidence appears. It is related to
multiple-instance learning~\citep{ilse2018attentionmil}, where a collection of
instances receives one global label. However, standard multiple-instance
aggregation is often permutation invariant, whereas conversational evidence
is ordered and may accumulate across utterances. A suitable model must
therefore preserve local temporal structure while learning from coarse
conversation-level supervision.

\begin{figure}[t]
    \centering
    \includegraphics[width=\linewidth]{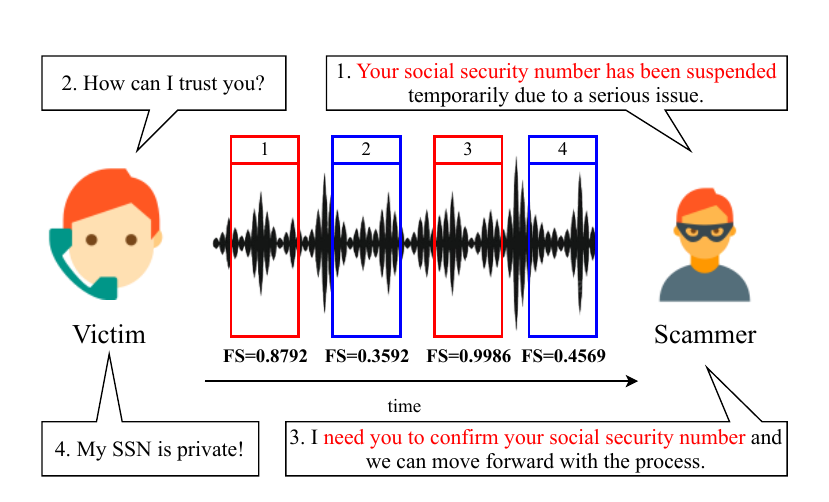}
    \caption{
        An SSN scam example in which high latent fraud scores occur only for
        selected utterances. The scores illustrate how estimated risk can
        vary across a conversation, but should not be interpreted as
        supervised fraud-span annotations.
    }
    \label{fig:scam_scenario}
\end{figure}

Figure~\ref{fig:scam_scenario} illustrates this challenge. A fraudulent call
may contain both suspicious and apparently benign exchanges, with informative
evidence appearing only at particular moments. A single post-call prediction
cannot describe how the estimated risk changes as the conversation unfolds.
We therefore introduce \textbf{StreamFraudNet}, a bounded-context model that
produces an initial fraud score from the first observed window and updates the
conversation-level prediction as additional audio arrives.

StreamFraudNet divides incoming audio into overlapping bounded-context
windows. A frozen Wav2Vec2 encoder
~\citep{baevski2020wav2vec2} extracts frame-level speech representations,
attention pooling forms chunk-level vectors, and a recurrent module captures
temporal dependencies within each observed window. A learned aggregator then
combines all window scores available so far into the current
conversation-level prediction. Because no temporal labels are provided, the
intermediate outputs are treated as latent risk scores rather than verified
localizations of fraudulent content.

Our contributions are as follows.

\begin{itemize}
    \item We formulate telecom fraud detection as weakly supervised
    incremental scoring from raw audio using only conversation-level labels.

    \item We propose StreamFraudNet, a bounded-context architecture with a
    frozen speech encoder, recurrent temporal modeling, and incremental
    aggregation. Its \(0.953\)M-parameter task head produces the first score
    after 10 seconds and updates it every 2 seconds.

    \item Through controlled multi-seed experiments, we show that recurrent
    temporal context is the principal architectural contributor.
    StreamFraudNet remains competitive with strong temporal baselines and
    operates faster than real time, while bounded windows primarily enable
    incremental operation rather than higher final-call accuracy.
\end{itemize}

%% file: sections/related_work.tex
\section{Related Work}
\label{sec:related}

\subsection{Metadata- and Graph-Based Fraud Detection}

Much telecom fraud research relies on Call Detail Records and interaction
metadata, including call duration, frequency, timing, and communication
structure
~\citep{prasad2020whoscalling,hu2024gatcobo,cao2024tfdgcl}.
Graph-based methods represent users, calls, or transactions as relational
structures and learn patterns associated with fraudulent behavior.
\citet{hu2024gatcobo} introduced cost-sensitive graph learning for telecom
fraud detection, while \citet{cao2024tfdgcl} combined graph contrastive
learning with adaptive augmentation.

Related transaction-fraud methods model class imbalance, neighborhood
structure, and temporal dependencies
~\citep{tian2024asagnn,tian2024spatiotemporal,xie2024stgn}.
Although these studies demonstrate the value of relational and temporal
information, they operate on metadata or transaction graphs rather than
spoken conversations. They therefore motivate temporal modeling broadly but
are not direct baselines for raw-audio fraud detection.

\subsection{Acoustic and Conversational Fraud Detection}

Earlier audio-based work focused on acoustic and spectro-temporal properties
of recorded calls. \citet{prasad2020whoscalling} combined audio and metadata
to characterize robocalls, while \citet{elizalde2021detection} evaluated
handcrafted acoustic descriptors and spectrogram-based classifiers for
robocall and voicemail-spam detection. These studies establish that call
audio contains discriminative information, but primarily address
recorded-message classification rather than fraud evidence that develops
throughout an interactive conversation.

A complementary direction models conversational semantics.
\citet{shen2025where} examined the capabilities and limitations of
LLM-based phone-scam detection, while \citet{shen2025rightmoment} studied
real-time warnings from partially observed call transcripts. Such systems
can reason over linguistic content as a call unfolds, but depend on
automatic speech recognition and operate mainly in the textual domain.

More recent work uses end-to-end audio--language models.
\citet{ma2025teleantifraud} introduced TeleAntiFraud-28K and adapted
Qwen2-Audio for telecom fraud classification and reasoning.
\citet{wang2026safeqaq} subsequently proposed audio--text fraud detection
with reinforcement learning and dynamic risk assessment. These approaches
use high-capacity audio--language models and reasoning-oriented supervision.
Our setting is complementary: we study bounded-context fraud scoring
directly from raw audio using a compact trainable head and only binary
conversation-level labels.

\subsection{Weakly Supervised Temporal Modeling}

Our supervision setting is related to multiple-instance learning, where a
set of instances receives a single bag-level label.
\citet{ilse2018attentionmil} introduced attention-based aggregation that
learns instance importance from global supervision. Standard
multiple-instance models are typically permutation invariant
~\citep{pmlr-v97-lee19d,9423289}, however, and do not explicitly represent
the order in which conversational evidence appears.

Weakly supervised sound event detection similarly learns frame- or
segment-level scores from clip-level labels. \citet{miyazaki2020weakly}
applied self-attention in this setting, while
\citet{deshmukh2021weakly} introduced auxiliary self-supervised objectives
for learning from coarse annotations. Sound events are often temporally
localized acoustic phenomena
~\citep{Kotus2014,politis2021dcase}, whereas telecom fraud may depend on
evidence distributed across multiple utterances. StreamFraudNet therefore
combines conversation-level supervision with ordered, bounded-context
temporal modeling. Because timestamped fraud annotations are unavailable,
its window outputs are treated as latent risk scores rather than verified
event localizations.

%% file: sections/methodology.tex
\section{Proposed Methodology}
\label{sec:method}

\subsection{Problem Formulation}

Let \(\mathbf{x}=(x_1,\ldots,x_C)\) denote a telephone conversation
divided into \(C\) fixed-duration audio chunks, where each
\(x_c \in \mathbb{R}^{T}\) contains \(T\) waveform samples. Each
conversation has a binary label \(y \in \{0,1\}\), indicating a legitimate
or fraudulent call. During training, only this conversation-level label is
available; no chunk-, utterance-, or timestamp-level annotations are
provided.

We formulate the task as weakly supervised incremental fraud scoring. At
time \(t\), let \(C_t\) denote the number of chunks observed so far. The
observed audio is divided into overlapping windows of \(k\) chunks with
stride \(s\). The \(i\)-th complete window is
\begin{equation}
\mathbf{w}_i
=
\left(
x_{is+1},
x_{is+2},
\ldots,
x_{is+k}
\right).
\label{eq:window}
\end{equation}
The valid indices are
\(i \in \{0,\ldots,N_t-1\}\), where the number of complete windows
available at time \(t\) is
\begin{equation}
N_t
=
\left\lfloor
\frac{C_t-k}{s}
\right\rfloor
+
1,
\label{eq:num_observed_windows}
\end{equation}
where \(C_t \geq k\). At the end of the conversation,
\(C_t=C\), and the total number of windows is
\(N=\lfloor(C-k)/s\rfloor+1\).

A window detector \(f_{\theta}\) maps each observed window to a latent fraud
score:
\begin{equation}
\hat{y}_i
=
f_{\theta}(\mathbf{w}_i)
\in [0,1].
\label{eq:local_score}
\end{equation}

The scores available at time \(t\) form the ordered sequence \(\hat{\mathbf{y}}^{(t)}
=(\hat{y}_0,\ldots,\hat{y}_{N_t-1})\)
are combined into the current prediction
\(\hat{y}_{\mathrm{global}}^{(t)}
=g_{\psi}(\hat{\mathbf{y}}^{(t)})\). As new chunks arrive, additional windows become available and
\(\hat{y}_{\mathrm{global}}^{(t)}\) is updated. Because temporal
annotations are unavailable, the individual \(\hat{y}_i\) values are
treated as latent risk scores rather than supervised localizations of
fraudulent content.

\subsection{Model Architecture}

\begin{figure}[t]
    \centering
    \includegraphics[width=\linewidth]{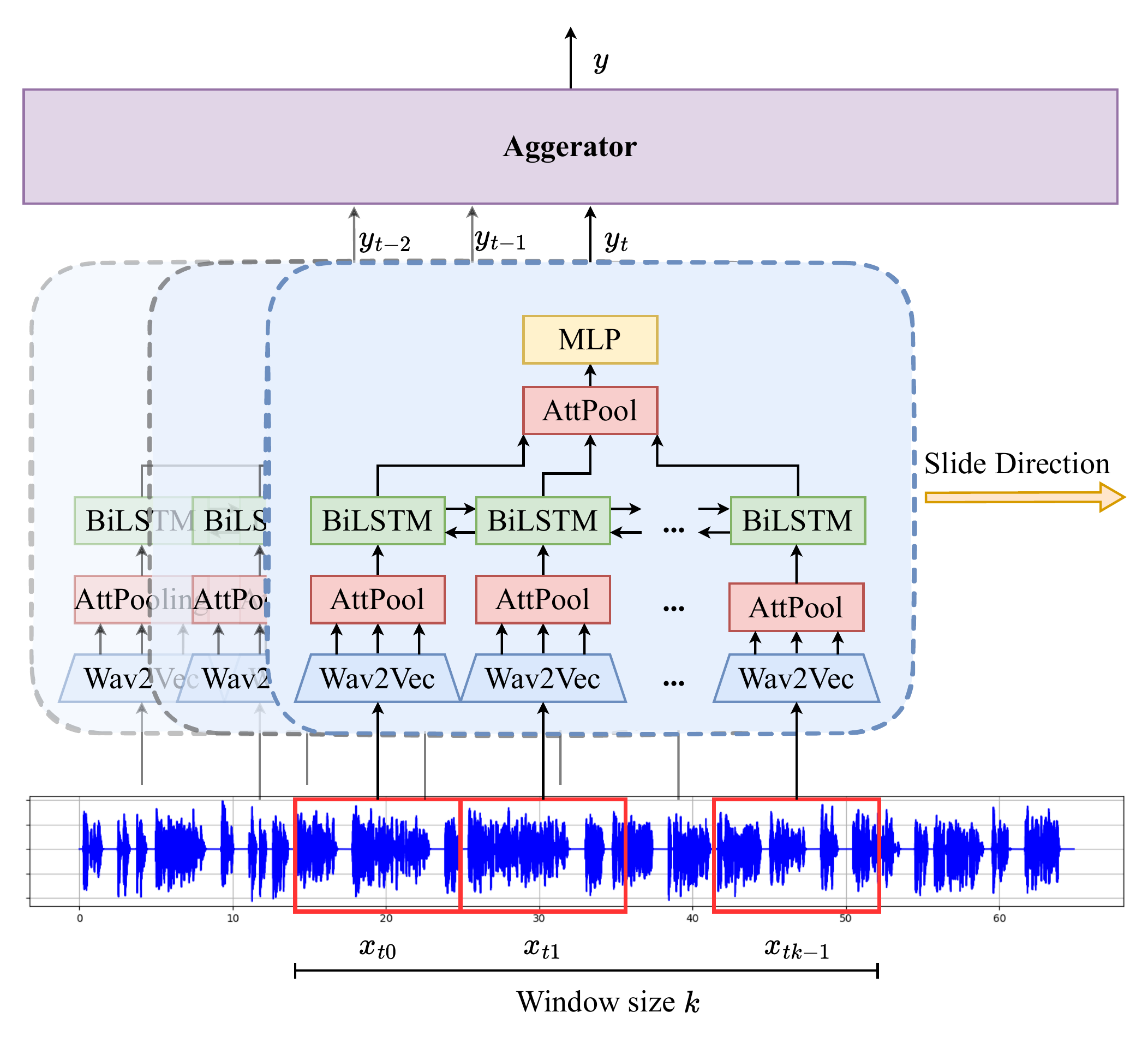}
    \caption{
        Overview of StreamFraudNet.
StreamFraudNet processes overlapping \(k\)-chunk windows using a frozen
Wav2Vec2 encoder, attention pooling, and a BiLSTM. The resulting latent
window scores are incrementally aggregated into the current call-level score.
    }
    \label{fig:architecture}
\end{figure}

StreamFraudNet contains a bounded-context window detector and an incremental
aggregator. The detector combines a frozen speech encoder, attention pooling,
a BiLSTM, and a window classifier, while the aggregator combines all window
scores observed so far. Figure~\ref{fig:architecture} summarizes the model.

\subsubsection{Bounded-Context Window Detector}

For the \(i\)-th window, let \(x_{i,j}=x_{is+j}\) denote its \(j\)-th
chunk, with \(j \in \{1,\ldots,k\}\). A frozen Wav2Vec2 encoder
\(\mathcal{E}_{\phi}\) extracts frame-level representations:

\begin{equation}
\mathbf{F}_{i,j}
=
\mathcal{E}_{\phi}(x_{i,j})
\in
\mathbb{R}^{L \times D},
\label{eq:wav2vec}
\end{equation}
where \(L\) is the number of encoder frames and \(D\) is the representation
dimension. Wav2Vec2 provides self-supervised speech representations without
requiring transcripts at inference time~\citep{baevski2020wav2vec2}. We
freeze the encoder to reduce trainable parameters and training-time memory.

Frame-level attention pooling converts the variable-length encoder output
into a fixed-dimensional chunk representation. The normalized attention
weight for frame \(\ell\) is
\begin{equation}
a_{i,j,\ell}
=
\frac{
    \exp\left(
        \mathbf{u}^{\top}\mathbf{F}_{i,j,\ell}
    \right)
}{
    \sum_{m=1}^{L}
    \exp\left(
        \mathbf{u}^{\top}\mathbf{F}_{i,j,m}
    \right)
}
\label{eq:frame_attention}
\end{equation}
where \(\mathbf{u} \in \mathbb{R}^{D}\) is a trainable attention vector.
The pooled chunk representation is
\begin{equation}
\mathbf{c}_{i,j}
=
\sum_{\ell=1}^{L}
a_{i,j,\ell}
\mathbf{F}_{i,j,\ell}.
\label{eq:chunk_representation}
\end{equation}

The \(k\) pooled chunk representations form an ordered window sequence:
\begin{equation}
\mathbf{C}_i
=
\left(
\mathbf{c}_{i,1},
\ldots,
\mathbf{c}_{i,k}
\right)
\in
\mathbb{R}^{k \times D}.
\label{eq:window_sequence}
\end{equation}

A bidirectional LSTM models temporal dependencies among chunks within the
current window:
\begin{equation}
\mathbf{H}_i
=
\operatorname{BiLSTM}(\mathbf{C}_i)
\in
\mathbb{R}^{k \times 2H}
\label{eq:bilstm}
\end{equation}
where \(H\) is the hidden dimension in each direction. Bidirectionality is
restricted to the current observed window. The detector may use all chunks
within \(\mathbf{w}_i\), but it cannot access audio from future windows.

A second attention layer summarizes the contextualized chunk
representations. The normalized weight assigned to chunk \(j\) is
\begin{equation}
b_{i,j}
=
\frac{
    \exp\left(
        \mathbf{v}^{\top}\mathbf{H}_{i,j}
    \right)
}{
    \sum_{r=1}^{k}
    \exp\left(
        \mathbf{v}^{\top}\mathbf{H}_{i,r}
    \right)
}
\label{eq:window_attention}
\end{equation}
where \(\mathbf{v} \in \mathbb{R}^{2H}\) is a trainable attention vector.
The resulting window representation is
\begin{equation}
\mathbf{h}_i
=
\sum_{j=1}^{k}
b_{i,j}
\mathbf{H}_{i,j}.
\label{eq:window_representation}
\end{equation}

A multilayer perceptron maps the window representation to a latent fraud
score:
\begin{equation}
\hat{y}_i
=
\sigma
\left(
\operatorname{MLP}(\mathbf{h}_i)
\right).
\label{eq:window_prediction}
\end{equation}

Applying the detector to all complete windows available at time \(t\) yields
the ordered score sequence
\(\hat{\mathbf{y}}^{(t)}
=(\hat{y}_0,\ldots,\hat{y}_{N_t-1})\),
which is passed to the incremental aggregator.

\subsubsection{Incremental Conversation-Level Aggregator}

At each inference step, the aggregator combines all window scores observed
up to time \(t\). The normalized aggregation weight for window \(i\) is
\begin{equation}
\gamma_i^{(t)}
=
\frac{
    \exp\left(
        q\hat{y}_i
    \right)
}{
    \sum_{r=0}^{N_t-1}
    \exp\left(
        q\hat{y}_r
    \right)
}
\label{eq:aggregation_attention}
\end{equation}
where \(q \in \mathbb{R}\) is a trainable scalar. The current
conversation-level prediction is
\begin{equation}
\hat{y}_{\mathrm{global}}^{(t)}
=
\sum_{i=0}^{N_t-1}
\gamma_i^{(t)}
\hat{y}_i.
\label{eq:aggregated_prediction}
\end{equation}

When a new complete window becomes available, its score is appended to
\(\hat{\mathbf{y}}^{(t)}\). The aggregation weights and
conversation-level prediction are then recomputed using all windows
observed so far. The aggregator therefore produces updated fraud scores
during the conversation and does not require the call to end. The score-dependent weighting allows different windows to contribute
unequally to the current prediction.

\subsection{Weakly Supervised Training}

During training, all complete windows from a conversation are processed.
Let \(t_{\mathrm{end}}\) denote the end of the conversation, such that
\(N_{t_{\mathrm{end}}}=N\). The final training prediction is
\(\hat{y}_{\mathrm{global}}
=
\hat{y}_{\mathrm{global}}^{(t_{\mathrm{end}})}\).

All trainable components are optimized jointly using the
conversation-level binary cross-entropy objective:
\begin{equation}
\mathcal{L}_{\mathrm{BCE}}
=
-y
\log
\left(
\hat{y}_{\mathrm{global}}
\right)
-
(1-y)
\log
\left(
1-\hat{y}_{\mathrm{global}}
\right).
\label{eq:bce}
\end{equation}

Gradients propagate through the conversation-level aggregator, window
classifier, BiLSTM, and attention modules, while the pretrained speech
encoder remains fixed. No auxiliary objective is applied to individual
window scores, intermediate conversation prefixes, or temporal locations.

The model is therefore trained to discriminate complete conversations
using only conversation-level labels. Intermediate predictions are
available because the same aggregator can be applied to any observed
prefix, but these predictions are not explicitly optimized for early
detection. Likewise, the training objective does not guarantee that an
individual window score corresponds to a temporally localized fraud event.

\subsection{Incremental Inference and Efficiency}

After the first \(k\) chunks have been observed, StreamFraudNet produces one
window score and its first conversation-level prediction. The prediction is
then updated whenever \(s\) additional chunks arrive. For chunk duration
\(\Delta\), the initial observation duration is \(k\Delta\), and the update
interval is \(s\Delta\).

Under the default configuration \(k=5\), \(s=1\), and
\(\Delta=2\) seconds, the first prediction is produced after 10 seconds of
audio and updated every 2 seconds thereafter. At each update, the
aggregator recomputes the conversation-level score using all complete
windows observed up to that point.

StreamFraudNet is incremental at the window level rather than strictly
causal at the frame level. The BiLSTM uses both temporal directions within
a complete observed window, but neither the window detector nor the
aggregator can access future, unheard audio.

We measure end-to-end computational throughput using the real-time factor

\begin{equation}
\mathrm{RTF}
=
\frac{
t_{\mathrm{processing}}
}{
t_{\mathrm{audio}}
},
\label{eq:rtf}
\end{equation}
where \(t_{\mathrm{processing}}\) is the time required to decode,
preprocess, encode, and classify the audio, and \(t_{\mathrm{audio}}\) is
its duration. An RTF below \(1\) indicates that the complete inference
pipeline processes audio faster than it is received.

RTF measures computational throughput rather than fraud-onset detection
delay. Fraud-onset delay also depends on when discriminative evidence first
appears and cannot be measured directly without timestamped fraud
annotations.

%% file: sections/experiment_result.tex
\section{Experimental Evaluation}
\label{sec:experiments}

\subsection{Datasets and Experimental Setup}

\begin{figure}[t]
    \centering
    \includegraphics[width=0.85\linewidth]
    {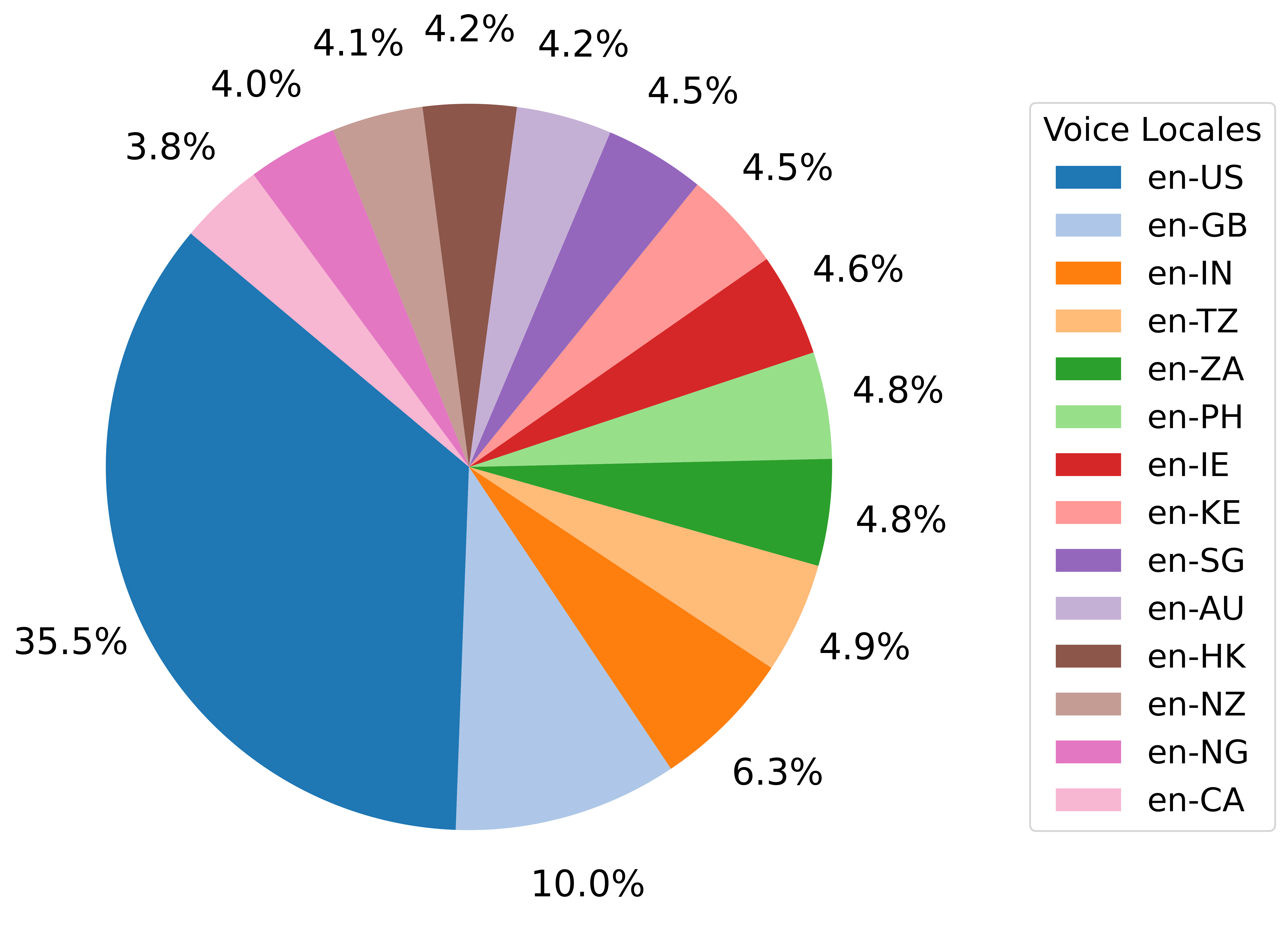}
    \caption{
        Distribution of accent locales across the 47 English-speaking voices
        used to synthesize the controlled English benchmark.
    }
    \label{fig:voices}
\end{figure}

\paragraph{Datasets.}
We evaluate on two conversation-level fraud benchmarks. The synthetic
English benchmark is constructed from the Scam Dialogue
Dataset\footnote{\url{https://huggingface.co/datasets/BothBosu/scam-dialogue}}. Each dialogue is synthesized with Edge-TTS
using two voices sampled from a pool of 47 English speakers (Figure~\ref{fig:voices}). The dataset
contains 1,600 balanced conversations, divided into 1,280 training and
320 held-out test examples. Each dialogue is synthesized with Edge-TTS using two voices sampled from
47 English-speaking speakers; their accent-locale distribution is reported
in Figure~\ref{fig:voices}.

We additionally evaluate on TeleAntiFraud-28K, a Mandarin audio-text
benchmark containing 28,511 telephone conversations with conversation-level
fraud labels~\cite{ma2025teleantifraud}. We follow its official split of
21,490 training and 7,021 test conversations. Table~\ref{tab:dataset_statistics}
summarizes both datasets.

\begin{table}[t]
    \centering
    \small
    \caption{Dataset statistics.}
    \label{tab:dataset_statistics}
    \resizebox{\columnwidth}{!}{
        \begin{tabular}{llrrr}
            \toprule
            \textbf{Dataset}
            & \textbf{Split}
            & \textbf{Total}
            & \textbf{Fraud}
            & \textbf{Legitimate} \\
            \midrule
            \multirow{2}{*}{Synthetic}
            & Train & 1,280 & 640 & 640 \\
            & Test  & 320   & 160 & 160 \\
            \midrule
            \multirow{2}{*}{TeleAntiFraud-28K}
            & Train & 21,490 & 9,950 & 11,540 \\
            & Test  & 7,021  & 3,697 & 3,324 \\
            \bottomrule
        \end{tabular}
    }
\end{table}

\paragraph{Model configuration.}
All recordings are converted to mono and resampled to 16~kHz. For the
synthetic benchmark, we use a frozen Wav2Vec2-Base encoder
~\cite{baevski2020wav2vec2}. Conversations are divided into 2-second chunks
and truncated or padded to 47 chunks. StreamFraudNet uses five-chunk windows,
stride one, a single-layer BiLSTM with 128 hidden units per direction, and
dropout 0.2. The resulting local context spans 10 seconds, with score updates
every 2 seconds.

For TeleAntiFraud-28K, we use a frozen Mandarin Wav2Vec2-Large-XLSR~\cite{conneau21_interspeech}
encoder, retain at most 30 chunks, and set the stride to two. The
task-specific architecture is otherwise unchanged.

\paragraph{Training and evaluation.}
Models are trained with AdamW and binary cross-entropy for up to 25 epochs,
using cosine learning-rate decay and early stopping on validation F1.
Learning rates are selected separately under the same validation protocol.
Experiments use NVIDIA GPUs with 40\,GB memory and a 12-core CPU;
end-to-end inference results are reported in
Section~\ref{sec:efficiency}.

Controlled experiments use seeds 13, 37, and 73, with mean and sample
standard deviation reported across runs. Three-seed ensembles average
prediction scores and use validation-selected thresholds. We estimate
confidence intervals using 10,000 label-stratified bootstrap samples and
apply paired bootstrap tests for ROC--AUC and F1, exact McNemar tests for
accuracy, and Holm correction within each test family.

\paragraph{Baselines.}
We compare StreamFraudNet with three groups of baselines. Acoustic controls
use call duration or log-Mel features with linear and temporal CNN
classifiers, testing whether simple signal-level cues are sufficient.
Representation baselines apply mean pooling, Transformer attention, or
BiLSTM attention to frozen Wav2Vec2 features, isolating the contribution of
temporal modeling. We additionally include Wav2Vec2-Large as an encoder
capacity control and WavLM-Base+ with the matched BiLSTM-attention head
~\citep{chen2022wavlm} to assess sensitivity to the pretrained speech
encoder.

\subsection{Conversation-Level Classification}

\begin{table*}[t]
    \centering
    \small
    \resizebox{\textwidth}{!}{%
    \begin{threeparttable}
        \caption{
            Conversation-level performance on the synthetic test set.
            Results are mean \(\pm\) sample standard deviation over three
            seeds.
        }
        \label{tab:main_results}
        \begin{tabular}{@{}lccccc@{}}
            \toprule
            \textbf{Model}
            & \textbf{Head params.}
            & \textbf{ROC--AUC}
            & \textbf{F1}
            & \textbf{Ens. AUC}
            & \textbf{95\% CI} \\
            \midrule
            \multicolumn{6}{@{}l}{\textit{Acoustic controls}} \\
            \addlinespace[1pt]
            Duration mean--linear\(^{\dagger}\)
            & 2
            & \(0.6236 \pm 0.4068\)
            & \(0.7235 \pm 0.0492\)
            & 0.8585
            & \([0.8174, 0.8958]\) \\
            Log-Mel mean--linear\(^{\dagger}\)
            & 161
            & \(0.6355 \pm 0.0168\)
            & \(0.7019 \pm 0.0058\)
            & 0.6402
            & \([0.5778, 0.7002]\) \\
            Log-Mel temporal CNN\(^{\dagger}\)
            & 127K
            & \(0.8718 \pm 0.0158\)
            & \(0.7943 \pm 0.0149\)
            & 0.8795
            & \([0.8406, 0.9143]\) \\
            \addlinespace[4pt]
            \multicolumn{6}{@{}l}{\textit{Wav2Vec2 controls}} \\
            \addlinespace[1pt]
            Wav2Vec2 mean--linear\(^{\dagger}\)
            & 769
            & \(0.7934 \pm 0.0146\)
            & \(0.7295 \pm 0.0235\)
            & 0.8020
            & \([0.7544, 0.8477]\) \\
            Wav2Vec2-Large mean--linear\(^{\dagger}\)
            & 1.03K
            & \(0.7244 \pm 0.0401\)
            & \(0.7189 \pm 0.0106\)
            & 0.6830
            & \([0.6221, 0.7407]\) \\
            Wav2Vec2 Transformer-attention
            & 7.98M
            & \(0.9784 \pm 0.0116\)
            & \(0.9525 \pm 0.0172\)
            & 0.9818
            & \([0.9654, 0.9944]\) \\
            Wav2Vec2 BiLSTM-attention
            & 0.953M
            & \(\mathbf{0.9990 \pm 0.0008}\)
            & \(\underline{0.9833 \pm 0.0066}\)
            & \(\underline{0.9992}\)
            & \([0.9975, 1.0000]\) \\
            \addlinespace[4pt]
            \multicolumn{6}{@{}l}{
                \textit{Alternative encoder and proposed model}
            } \\
            \addlinespace[1pt]
            WavLM BiLSTM-attention\(^{\ddagger}\)
            & 0.953M
            & \(\mathbf{0.9990 \pm 0.0009}\)
            & \(\mathbf{0.9834 \pm 0.0049}\)
            & \(\mathbf{0.9998}\)
            & \(\mathbf{[0.9994, 1.0000]}\) \\
            \textbf{StreamFraudNet}
            & 0.953M
            & \(\underline{0.9953 \pm 0.0022}\)
            & \(0.9656 \pm 0.0136\)
            & 0.9962
            & \([0.9916, 0.9994]\) \\
            \bottomrule
        \end{tabular}
        \begin{tablenotes}[flushleft]
            \footnotesize
            \item Best values are shown in bold and second-best distinct
            values are underlined. Head parameters exclude the frozen speech
            encoder.
            \item[\(\dagger\)] Significantly lower ROC--AUC and F1 than
            StreamFraudNet after Holm correction.
            \item[\(\ddagger\)] Significantly higher ensemble ROC--AUC than
            StreamFraudNet after Holm correction.
        \end{tablenotes}
    \end{threeparttable}%
    }
\end{table*}

Table~\ref{tab:main_results} compares StreamFraudNet with acoustic,
representation, and temporal baselines. StreamFraudNet significantly
outperforms the acoustic and mean-pooling controls in both ROC--AUC and F1
(\(p_{\mathrm{Holm}}=0.0044\)). It also uses approximately 0.95 million
task-head parameters, compared with 7.98 million for Wav2Vec2
Transformer-attention, while obtaining higher mean ROC--AUC and F1.
The difference between these two models, however, is not significant after
Holm correction~\cite{29def780-e117-38f0-8afb-edf384af3fad}.

StreamFraudNet has essentially the same task-head size as the Wav2Vec2 and
WavLM BiLSTM-attention baselines. Its difference from Wav2Vec2
BiLSTM-attention is not significant, whereas WavLM BiLSTM-attention obtains
a significantly higher ensemble ROC--AUC, with a difference of \(0.00363\)
(\(95\%\ \mathrm{CI}=[0.00059,0.00801]\),
\(p_{\mathrm{Holm}}=0.0128\)). These results show that StreamFraudNet
provides competitive performance with a compact task-specific head and
bounded-context incremental operation, although it does not achieve the
strongest offline classification result. Complete pairwise tests and secondary metrics are
reported in Appendix~\ref{app:additional_experiments}.

\subsection{Component Analysis}
\label{sec:component_analysis}

StreamFraudNet combines bounded windows, recurrent modeling, attention
pooling, and a nonlinear classifier. To identify which choices are essential,
we remove or simplify one component at a time under the same evaluation
protocol, and present the results at Table~\ref{tab:ablations}.

\begin{table*}[t]
    \centering
    \small
    \caption{
        Component analysis on the synthetic test set. Performance is reported
        as mean \(\pm\) sample standard deviation across three seeds. Paired
        effects are included only when significant after Holm correction;
        ``n.s.'' denotes no significant difference.
    }
    \label{tab:ablations}
    \resizebox{\textwidth}{!}{
        \begin{tabular}{@{}lccc@{}}
            \toprule
            \textbf{Configuration}
            & \textbf{ROC--AUC}
            & \textbf{F1}
            & \textbf{Paired effect relative to full model} \\
            \midrule
            Full StreamFraudNet
            & \(0.9953 \pm 0.0022\)
            & \(0.9656 \pm 0.0136\)
            & -- \\

            Global single window
            & \(0.9990 \pm 0.0008\)
            & \(0.9833 \pm 0.0066\)
            & n.s. \\

            No recurrent context
            & \(0.9651 \pm 0.0065\)
            & \(0.8970 \pm 0.0067\)
            & \(\Delta\mathrm{AUC}=-0.02797\),
              \(\Delta\mathrm{F1}=-0.07637\);
              \(p_{\mathrm{Holm}}=0.0044\) \\

            Unidirectional LSTM
            & \(0.9951 \pm 0.0009\)
            & \(0.9694 \pm 0.0078\)
            & n.s. \\

            Local attention replaced by mean
            & \(0.9962 \pm 0.0015\)
            & \(0.9721 \pm 0.0028\)
            & n.s. \\

            Conversation attention replaced by mean
            & \(0.9939 \pm 0.0020\)
            & \(0.9560 \pm 0.0065\)
            & n.s. \\

            Nonlinear head replaced by linear
            & \(0.9910 \pm 0.0046\)
            & \(0.9503 \pm 0.0164\)
            & n.s. \\
            \bottomrule
        \end{tabular}
    }
\end{table*}

Removing recurrent context causes the only significant degradation,
reducing ROC--AUC by \(0.02797\) and F1 by \(0.07637\)
(\(p_{\mathrm{Holm}}=0.0044\); Table~\ref{tab:ablations}).
Bidirectionality, attention pooling, and the nonlinear head provide no
individually significant gain. The global single-window variant is
numerically stronger but not significantly different, indicating that
bounded windows primarily enable frequent updates rather than better
offline classification.

\subsection{Speech Encoder Training Policy}
\label{sec:encoder_policy}

Freezing the speech encoder reduces training cost but may limit adaptation
to fraud-specific cues. We therefore compare frozen, partially fine-tuned,
and fully fine-tuned encoders in Table~\ref{tab:encoder_policy}.

\begin{table*}[t]
    \centering
    \small
    \setlength{\tabcolsep}{14pt} % Điều chỉnh nếu cần (3pt--5pt)
    \caption{
        Effect of the speech-encoder training policy. Performance is reported
        as mean \(\pm\) sample standard deviation across three seeds. Peak
        memory denotes training-time GPU memory.
    }
    \label{tab:encoder_policy}

    \begin{tabular}{@{}lrrrr@{}}
        \toprule
        \textbf{Encoder policy}
        & \textbf{Trainable params.}
        & \textbf{Peak memory}
        & \textbf{ROC--AUC}
        & \textbf{F1} \\
        \midrule
        Frozen
        & 0.953M
        & 0.681 GiB
        & \(0.9916 \pm 0.0067\)
        & \(0.9591 \pm 0.0108\) \\
        Fine-tune final two layers
        & 15.129M
        & 2.586 GiB
        & \(0.9920 \pm 0.0072\)
        & \(0.9565 \pm 0.0219\) \\
        Full fine-tuning, LR \(10^{-6}\)
        & 95.325M
        & 3.809 GiB
        & \(0.9842 \pm 0.0038\)
        & \(0.9431 \pm 0.0091\) \\
        Full fine-tuning, LR \(10^{-5}\)
        & 95.325M
        & 3.809 GiB
        & \(0.8618 \pm 0.2349\)
        & \(0.8834 \pm 0.1668\) \\
        \bottomrule
    \end{tabular}
\end{table*}

Fine-tuning the final two layers provides no performance gain while
increasing peak memory from 0.681 to 2.586~GiB. Full fine-tuning requires
substantially more memory and is sensitive to the learning rate: \(10^{-5}\)
is unstable, while \(10^{-6}\) remains below the frozen configuration.

The frozen ensemble exceeds stable full fine-tuning by 0.00949 ROC--AUC
(\(95\%\ \mathrm{CI}=[0.00246,0.01809]\), \(p=0.0034\)). Freezing is
therefore justified by its stronger stability--cost trade-off, although the
encoder remains part of the inference pipeline.

\subsection{Hyperparameter Robustness}
\label{sec:hyperparameter_robustness}

Window size and stride control the amount of local context and the update
frequency, while hidden dimension and dropout affect capacity and
regularization. We vary each factor independently to determine whether the
results depend on a narrow configuration
(Table~\ref{tab:hyperparameter_sensitivity}).

\begin{table}[t]
    \centering
    \small
    \caption{
        Hyperparameter sensitivity on the synthetic test set. Values are
        mean \(\pm\) sample standard deviation across three seeds. No tested
        variant differs significantly from the default after Holm correction.
    }
    \label{tab:hyperparameter_sensitivity}
    \resizebox{\columnwidth}{!}{
        \begin{tabular}{@{}lcc@{}}
            \toprule
            \textbf{Configuration}
            & \textbf{ROC--AUC}
            & \textbf{F1} \\
            \midrule
            \textbf{Default: \(W=5,S=1,H=128,p=0.2\)}
            & \(0.9953 \pm 0.0022\)
            & \(0.9656 \pm 0.0136\) \\
            \midrule
            \(W=3\)
            & \(0.9891 \pm 0.0014\)
            & \(0.9415 \pm 0.0148\) \\

            \(W=7\)
            & \(0.9904 \pm 0.0060\)
            & \(0.9444 \pm 0.0185\) \\
            \midrule
            \(S=2\)
            & \(0.9928 \pm 0.0011\)
            & \(0.9609 \pm 0.0025\) \\

            \(S=3\)
            & \(0.9967 \pm 0.0006\)
            & \(\textbf{0.9771} \pm 0.0048\) \\
            \midrule
            \(H=64\)
            & \(0.9952 \pm 0.0020\)
            & \(0.9704 \pm 0.0086\) \\

            \(H=256\)
            & \(\textbf{0.9958} \pm 0.0022\)
            & \(0.9664 \pm 0.0104\) \\
            \midrule
            \(p=0\)
            & \(0.9952 \pm 0.0027\)
            & \(0.9696 \pm 0.0020\) \\

            \(p=0.4\)
            & \(0.9947 \pm 0.0020\)
            & \(0.9697 \pm 0.0149\) \\
            \bottomrule
        \end{tabular}
    }
\end{table}

No tested variant differs significantly from the default after Holm
correction. Performance is particularly stable across hidden dimensions and
dropout values, while temporal settings produce larger numerical variation. Stride three gives the highest mean scores but updates only every 6 seconds.
We retain stride one because it provides 2-second updates while remaining
statistically competitive, prioritizing responsiveness over the strongest
offline mean.

\subsection{Incremental Prediction}
\label{sec:incremental_prediction}

Final-call metrics do not show whether useful predictions are available
during an ongoing conversation. We therefore evaluate progressively longer
audio prefixes, beginning at 10 seconds when the first complete window is
available (Table~\ref{tab:partial_results}).

\begin{table}[t]
    \centering
    \small
    \caption{
        Performance as a function of observed audio duration. Values are
        mean \(\pm\) sample standard deviation across three seeds.
    }
    \label{tab:partial_results}
    \resizebox{\columnwidth}{!}{
        \begin{tabular}{@{}rccc@{}}
            \toprule
            \textbf{Audio}
            & \textbf{SFN AUC}
            & \textbf{SFN F1}
            & \textbf{Global BiLSTM AUC} \\
            \midrule
            10 s
            & \(0.9532 \pm 0.0178\)
            & \(0.8645 \pm 0.0352\)
            & \(0.9682 \pm 0.0323\) \\

            20 s
            & \(0.9842 \pm 0.0060\)
            & \(0.9336 \pm 0.0087\)
            & \(0.9911 \pm 0.0090\) \\

            40 s
            & \(0.9895 \pm 0.0035\)
            & \(0.9468 \pm 0.0094\)
            & \(0.9970 \pm 0.0013\) \\

            60 s
            & \(0.9959 \pm 0.0018\)
            & \(0.9658 \pm 0.0113\)
            & \(0.9989 \pm 0.0007\) \\
            \bottomrule
        \end{tabular}
    }
\end{table}

StreamFraudNet reaches a ROC--AUC of \(0.9532 \pm 0.0178\) after 10 seconds
and \(0.9842 \pm 0.0060\) after 20 seconds, showing that useful predictions
can be produced before the call ends.

The global BiLSTM remains numerically stronger at every prefix. Thus, the
results support partial-conversation fraud scoring, but not a distinct
early-detection advantage from bounded-window processing alone.

\subsection{Label Efficiency}
\label{sec:label_efficiency}

To test whether the model remains effective with limited coarse supervision,
we train StreamFraudNet and the matched global BiLSTM using progressively
smaller subsets of the conversation-level labels. Table~\ref{tab:label_efficiency}
summarizes the results; full mean and standard deviation values are provided
in Appendix~\ref{app:label_efficiency}.

\begin{table}[!t]
    \centering
    \small
    \setlength{\tabcolsep}{15pt} % mặc định khoảng 6pt
    \caption{
        Label efficiency. Entries report mean ROC--AUC/F1 across three seeds.
    }
    \label{tab:label_efficiency}

    \begin{tabular}{@{}rcc@{}}
        \toprule
        \textbf{Labels}
        & \textbf{StreamFraudNet}
        & \textbf{Global BiLSTM} \\
        \midrule
        10\%  & \(0.7992/0.7316\) & \(0.8605/0.7872\) \\
        25\%  & \(0.8918/0.7960\) & \(0.9593/0.9152\) \\
        50\%  & \(0.9739/0.9210\) & \(0.9993/0.9845\) \\
        100\% & \(0.9953/0.9656\) & \(0.9990/0.9833\) \\
        \bottomrule
    \end{tabular}
\end{table}
Performance improves consistently with additional labels, and StreamFraudNet
reaches a ROC--AUC of 0.9739 using half of the training set. The global
BiLSTM remains stronger at every label budget, indicating that StreamFraudNet
can learn from reduced conversation-level supervision but is not more
label-efficient than the matched global model.

\subsection{External Evaluation}
\label{sec:external_evaluation}

The controlled English benchmark uses balanced synthetic speech. To assess
performance beyond this setting, we evaluate on the larger Mandarin
TeleAntiFraud-28K benchmark~\cite{ma2025teleantifraud}. Published
Qwen2-Audio results provide contextual reference, although the systems use
different modalities and supervision (Table~\ref{tab:teleanti}).

\begin{table}[t]
    \centering
    \small
    \setlength{\tabcolsep}{7pt}
    \caption{
        Results on TeleAntiFraud-28K. Qwen2-Audio results are taken from
        the original publication. Best results are shown in bold.
    }
    \label{tab:teleanti}
    \setlength{\tabcolsep}{15pt}
    \begin{tabular}{@{}lcc@{}}
        \toprule
        \textbf{Model}
        & \textbf{Accuracy}
        & \textbf{F1} \\
        \midrule
        Qwen2-Audio Base
        & 0.6183
        & 0.5851 \\

        Qwen2-Audio ASR-text
        & 0.7127
        & 0.7127 \\

        Qwen2-Audio No-Think
        & 0.6831
        & 0.6932 \\

        Qwen2-Audio Think
        & \textbf{0.8422}
        & \textbf{0.8478} \\

        \textbf{StreamFraudNet}
        & 0.7131
        & 0.7058 \\
        \bottomrule
    \end{tabular}
\end{table}

StreamFraudNet achieves 0.7131 accuracy and 0.7058 F1. Its performance is close to Qwen2-Audio ASR-text despite operating directly
on audio with only binary conversation labels. Qwen2-Audio Think remains
substantially stronger, indicating the benefit of larger audio-language
models and reasoning supervision.

These results support cross-language and cross-dataset applicability, but
should not be interpreted as a controlled model ranking. Full precision and
recall results are reported in Appendix~\ref{app:teleanti_full}.

\subsection{End-to-End Efficiency}
\label{sec:efficiency}

Incremental detection requires the complete pipeline, rather than only the
task head, to process audio faster than it arrives. We therefore measure
latency from raw WAV input through preprocessing, encoder inference,
classification, and aggregation
(Table~\ref{tab:hardware}).

\begin{table}[t]
    \centering
    \small
    \caption{End-to-end inference efficiency on server.}
    \label{tab:hardware}
    \resizebox{\columnwidth}{!}{
        \begin{tabular}{@{}lrrr@{}}
            \toprule
            \textbf{Platform}
            & \textbf{p50}
            & \textbf{p95}
            & \textbf{RTF} \\
            \midrule
            RTX PRO 6000
            & 290.8 ms
            & 320.7 ms
            & 0.00249 \\

            AMD EPYC 9555, 4 threads
            & 5.184 s
            & 5.535 s
            & 0.0440 \\
            \bottomrule
        \end{tabular}
    }
\end{table}

Both evaluated server configurations achieve an RTF below \(1\),
demonstrating faster-than-real-time end-to-end processing. These measurements
establish processing throughput rather than mobile deployability or
fraud-onset latency. Additional CPU-only results are reported in
Appendix~\ref{app:cpu_efficiency}. These measurements establish server-side throughput, not mobile deployment
or fraud-onset latency. Model footprint, memory use, and detailed timing
settings are provided in Appendix~\ref{app:efficiency_details}.

%% file: sections/conclusion.tex
\section{Conclusion}

We introduced StreamFraudNet for weakly supervised incremental fraud scoring
from raw telephone audio. The model combines a frozen speech encoder with
bounded-context recurrent modeling and requires only conversation-level
labels. Its main advantage is fixed-context, frequently updated scoring with
a compact task-specific head and modest training requirements. Future work
should explore latency-aware training and evaluate detection delay on
natural multilingual calls.

%% file: sections/appendix.tex
\appendix

\section{Additional Experimental Results}
\label{app:additional_experiments}

Unless otherwise stated, the following experiments use the synthetic English
benchmark, training seeds 13, 37, and 73, and the evaluation protocol in
Section~\ref{sec:experiments}.

\subsection{Secondary Metrics and Seed Variation}
\label{app:secondary_metrics}

Table~\ref{tab:app_secondary_metrics} complements the primary ROC--AUC and
F1 results with threshold-independent, thresholded, and correlation-based
metrics.

\begin{table}[t]
    \centering
    \small
    \caption{
        Additional StreamFraudNet results on the synthetic test set.
        Confidence intervals are estimated using 10,000 label-stratified
        dialogue bootstraps.
    }
    \label{tab:app_secondary_metrics}
    \resizebox{\columnwidth}{!}{
        \begin{tabular}{@{}lcc@{}}
            \toprule
            \textbf{Metric}
            & \textbf{Mean \(\pm\) SD}
            & \textbf{Ensemble [95\% CI]} \\
            \midrule
            ROC--AUC
            & \(0.9953 \pm 0.0022\)
            & \(0.9962\ [0.9916,0.9994]\) \\
            PR--AUC
            & \(0.9950 \pm 0.0023\)
            & \(0.9960\ [0.9911,0.9994]\) \\
            Accuracy
            & \(0.9656 \pm 0.0136\)
            & \(0.9719\ [0.9531,0.9875]\) \\
            F1
            & \(0.9656 \pm 0.0136\)
            & \(0.9716\ [0.9521,0.9875]\) \\
            MCC
            & \(0.9313 \pm 0.0272\)
            & \(0.9439\ [0.9064,0.9753]\) \\
            EER
            & \(0.0333 \pm 0.0130\)
            & \(0.0313\ [0.0125,0.0500]\) \\
            \bottomrule
        \end{tabular}
    }
\end{table}

The three individual runs are reported in
Table~\ref{tab:app_per_seed}. ROC--AUC varies little across seeds, whereas
F1 exhibits greater variation because it depends on the selected decision
threshold.

\begin{table}[t]
    \centering
    \small
    \caption{Per-seed StreamFraudNet performance.}
    \label{tab:app_per_seed}
    \setlength{\tabcolsep}{32pt}
    \begin{tabular}{@{}rcc@{}}
        \toprule
        \textbf{Seed}
        & \textbf{ROC--AUC}
        & \textbf{F1} \\
        \midrule
        13 & 0.9942 & 0.9565 \\
        37 & 0.9978 & 0.9813 \\
        73 & 0.9938 & 0.9590 \\
        \bottomrule
    \end{tabular}
\end{table}

\subsection{Complete Statistical Comparisons}
\label{app:pairwise_tests}

Table~\ref{tab:app_pairwise_tests} reports the corrected pairwise
comparisons underlying the significance markers in
Table~\ref{tab:main_results}. ``n.s.'' denotes a difference that is not
significant after Holm correction.

\begin{table*}[t]
    \centering
    \small
    \caption{
        Pairwise comparisons against StreamFraudNet. The direction refers
        to the ensemble ROC--AUC difference.
    }
    \label{tab:app_pairwise_tests}
    \resizebox{\textwidth}{!}{
        \begin{tabular}{@{}lccc@{}}
            \toprule
            \textbf{Baseline}
            & \textbf{Direction}
            & \textbf{ROC--AUC test}
            & \textbf{F1 test} \\
            \midrule
            Duration mean--linear
            & StreamFraudNet \(>\)
            & \(p_{\mathrm{Holm}}=0.0044\)
            & \(p_{\mathrm{Holm}}=0.0044\) \\

            Log-Mel mean--linear
            & StreamFraudNet \(>\)
            & \(p_{\mathrm{Holm}}=0.0044\)
            & \(p_{\mathrm{Holm}}=0.0044\) \\

            Log-Mel temporal CNN
            & StreamFraudNet \(>\)
            & \(p_{\mathrm{Holm}}=0.0044\)
            & \(p_{\mathrm{Holm}}=0.0044\) \\

            Wav2Vec2 mean--linear
            & StreamFraudNet \(>\)
            & \(p_{\mathrm{Holm}}=0.0044\)
            & \(p_{\mathrm{Holm}}=0.0044\) \\

            Wav2Vec2-Large mean--linear
            & StreamFraudNet \(>\)
            & \(p_{\mathrm{Holm}}=0.0044\)
            & \(p_{\mathrm{Holm}}=0.0044\) \\

            Wav2Vec2 Transformer-attention
            & StreamFraudNet \(>\)
            & \(p_{\mathrm{raw}}=0.0072,\ 
               p_{\mathrm{Holm}}=0.1008\)
            & n.s. \\

            Wav2Vec2 BiLSTM-attention
            & Baseline \(>\)
            & n.s.
            & n.s. \\

            WavLM BiLSTM-attention
            & Baseline \(>\)
            & \(\Delta=0.00363,\ 
               95\%\ \mathrm{CI}=[0.00059,0.00801],\
               p_{\mathrm{Holm}}=0.0128\)
            & n.s. \\
            \bottomrule
        \end{tabular}
    }
\end{table*}

For the component analysis, removing recurrent context decreases ROC--AUC
by 0.02797, with a 95\% confidence interval of
\([0.01398,0.04422]\), and decreases F1 by 0.07637, with an interval of
\([0.04264,0.11200]\). Both comparisons have
\(p_{\mathrm{Holm}}=0.0044\). No other component ablation remains
significant after correction.

\subsection{Encoder Fine-Tuning Diagnostics}
\label{app:encoder_diagnostics}

The aggregate encoder-policy comparison is reported in
Table~\ref{tab:encoder_policy}. Table~\ref{tab:app_encoder_seed_results}
provides the per-seed results needed to explain the high variance observed
under full fine-tuning with learning rate \(10^{-5}\).

\begin{table}[t]
    \centering
    \small
    \caption{Per-seed ROC--AUC under full encoder fine-tuning.}
    \label{tab:app_encoder_seed_results}
    \setlength{\tabcolsep}{16pt}
        \begin{tabular}{@{}rcc@{}}
            \toprule
            \textbf{Seed}
            & \textbf{Encoder LR \(10^{-5}\)}
            & \textbf{Encoder LR \(10^{-6}\)} \\
            \midrule
            13 & 0.59053 & 0.98209 \\
            37 & 0.99648 & 0.98199 \\
            73 & 0.99836 & 0.98857 \\
            \bottomrule
        \end{tabular}
    
\end{table}

The instability at learning rate \(10^{-5}\) is concentrated in seed 13.
Reducing the learning rate eliminates this failure, but the resulting
ensemble remains below the frozen encoder, as discussed in
Section~\ref{sec:encoder_policy}.

\subsection{Full TeleAntiFraud-28K Results}
\label{app:teleanti_full}

Table~\ref{tab:app_teleanti_full} supplements the accuracy and F1 values
reported in Section~\ref{sec:external_evaluation} with precision and recall.

\begin{table}[t]
    \centering
    \small
    \caption{
        Full TeleAntiFraud-28K results. Published Qwen2-Audio values are
        contextual rather than controlled comparisons.
    }
    \label{tab:app_teleanti_full}
    \resizebox{\columnwidth}{!}{
        \begin{tabular}{@{}lrrrrr@{}}
            \toprule
            \textbf{Model}
            & \textbf{Acc.}
            & \textbf{Prec.}
            & \textbf{Rec.}
            & \textbf{F1}
            & \textbf{AUC} \\
            \midrule
            Qwen2-Audio Base
            & 0.6183 & 0.6840 & 0.5112 & 0.5851 & -- \\
            Qwen2-Audio ASR-text
            & 0.7127 & 0.7639 & 0.6680 & 0.7127 & -- \\
            Qwen2-Audio No-Think
            & 0.6831 & 0.7404 & 0.6517 & 0.6932 & -- \\
            Qwen2-Audio Think
            & 0.8422 & 0.8615 & 0.8345 & 0.8478 & -- \\
            StreamFraudNet
            & 0.7131 & 0.8333 & 0.6122 & 0.7058 & 0.8100 \\
            \bottomrule
        \end{tabular}
    }
\end{table}

StreamFraudNet has higher precision but lower recall than Qwen2-Audio
ASR-text, explaining their similar accuracy but slightly different F1
scores.

\subsection{Model Footprint and Inference Protocol}
\label{app:efficiency_details}

The full model footprint and streaming configuration are reported in
Table~\ref{tab:app_model_footprint}. Although only 0.953 million task-head
parameters are trained, the frozen speech encoder remains part of the
deployed pipeline.

\begin{table}[t]
    \centering
    \small
    \caption{Model footprint and streaming configuration.}
    \label{tab:app_model_footprint}
    \setlength{\tabcolsep}{30pt}
        \begin{tabular}{@{}lc@{}}
            \toprule
            \textbf{Quantity} & \textbf{Value} \\
            \midrule
            Trainable task-head parameters & 0.953M \\
            Frozen encoder parameters & 94.372M \\
            Complete pipeline parameters & 95.325M \\
            Task checkpoint size & 10.93 MiB \\
            Full FP32 parameter storage & 363.6 MiB \\
            Local context duration & 10 s \\
            Score update interval & 2 s \\
            \bottomrule
        \end{tabular}
    
\end{table}

The end-to-end measurements in Table~\ref{tab:hardware} include raw audio
decoding, resampling, chunking, speech-encoder inference, task-head
inference, and final aggregation. They use batch size one, 10 warm-up runs,
and 100 timed conversations averaging 109.6 seconds, with audio capped at
120 seconds. These measurements quantify processing throughput on the
evaluated server hardware rather than mobile deployment or fraud-onset
detection latency.

\subsection{Label-Efficiency Details}
\label{app:label_efficiency}

The main paper summarizes performance under reduced supervision using
mean ROC--AUC/F1 pairs. Table~\ref{tab:app_label_efficiency} reports the
complete mean and sample standard deviation across the three training seeds.
The subsets contain 102, 256, 512, and 1,023 labeled conversations,
corresponding to 10\%, 25\%, 50\%, and 100\% of the available training data.

\begin{table*}[t]
    \centering
    \small
    \caption{
        Performance under reduced conversation-level supervision. Values
        are mean \(\pm\) sample standard deviation across three seeds.
    }
    \label{tab:app_label_efficiency}
    \resizebox{\textwidth}{!}{
        \begin{tabular}{@{}rcccc@{}}
            \toprule
            \textbf{Labels}
            & \textbf{StreamFraudNet ROC--AUC}
            & \textbf{StreamFraudNet F1}
            & \textbf{Global BiLSTM ROC--AUC}
            & \textbf{Global BiLSTM F1} \\
            \midrule
            10\%
            & \(0.7992 \pm 0.0105\)
            & \(0.7316 \pm 0.0100\)
            & \(0.8605 \pm 0.0579\)
            & \(0.7872 \pm 0.0571\) \\

            25\%
            & \(0.8918 \pm 0.0470\)
            & \(0.7960 \pm 0.0559\)
            & \(0.9593 \pm 0.0581\)
            & \(0.9152 \pm 0.0956\) \\

            50\%
            & \(0.9739 \pm 0.0157\)
            & \(0.9210 \pm 0.0404\)
            & \(0.9993 \pm 0.0001\)
            & \(0.9845 \pm 0.0030\) \\

            100\%
            & \(0.9953 \pm 0.0022\)
            & \(0.9656 \pm 0.0136\)
            & \(0.9990 \pm 0.0008\)
            & \(0.9833 \pm 0.0066\) \\
            \bottomrule
        \end{tabular}
    }
\end{table*}

Performance generally improves as additional conversation-level labels
become available. StreamFraudNet retains substantial discrimination with
half of the training data, reaching a mean ROC--AUC of 0.9739 and an F1
score of 0.9210. Its three-seed ensemble reaches 0.9853 ROC--AUC and
0.9536 F1 in this setting; relative to full-data training, the F1 reduction
is not significant after correction, whereas the ROC--AUC reduction is
significant.

The global BiLSTM remains numerically stronger at every evaluated label
budget. The experiment therefore demonstrates that StreamFraudNet can learn
from reduced conversation-level supervision, but does not establish greater
label efficiency than the matched global model.

\subsection{CPU-Only Inference Results}
\label{app:cpu_efficiency}

To assess computational efficiency under resource-constrained conditions, we sample 500 training instances per dataset (seed=42; 1,000 total) and evaluate all models CPU-only on commodity hardware (8 cores, 16GB RAM). We simulate real-time streaming by segmenting audio into 2-second chunks, using a 5-chunk (10-second) sliding window with stride 1. Real-time factor (RTF) is reported as the average ratio of inference time to audio duration per window, per dataset.

\begin{table}[ht]
\centering
\small
\setlength{\tabcolsep}{4pt}
\caption{Computational efficiency under CPU-only inference.}
\label{tab:compute}
\begin{tabular}{llcc}
\toprule
\textbf{Dataset} & \textbf{CPU Type} & \textbf{GFLOPs} & \textbf{RTF} \\
\midrule
\multirow{2}{*}{Synthetic} 
 & Apple M2        & 69.2496               & 0.0437             \\
 & Intel i5-13420H &  69.2496              &   0.1615           \\
\midrule
\multirow{2}{*}{TeleAntiFraud28K} 
 & Apple M2        & 178.7490                & 0.0835             \\
 & Intel i5-13420H & 178.7490                & 0.4168             \\
\bottomrule
\end{tabular}
\end{table}

As shown in Table~\ref{tab:compute}, the computational cost varies from 69.2 to 178.7 GFLOPs due to the use of different pretrained size of Wav2Vec encoders. Under CPU-only inference, the proposed model achieves real-time performance across all configurations, with RTFs ranging from 0.0437 to 0.4168 in a streaming setting. Notably, even the most computationally demanding configuration (178.7 GFLOPs) maintains an RTF well below 1, corresponding to approximately 2.5× faster-than-real-time processing. These results demonstrate that the model remains deployable in resource-constrained environments without GPU acceleration.

\section{Streaming Inference on Synthetic Fraudulent Conversations}

Tables~19 and 20 provide a closer, interpretability-oriented view of how StreamFraudNet distributes fraud evidence across sliding windows, complementing the aggregate ROC--AUC and F1 results reported in Section~4. Rather than treating the aggregator's output as a single opaque decision, these window-level traces let us inspect \emph{where} within the 10-second receptive field the local detector concentrates its score mass, and whether that concentration aligns with human-interpretable fraud cues. As illustrated in Figure.~\ref{fig:scam_scenario}, fraudulent intent is often expressed only at specific moments within an otherwise benign call, making call-level classification insufficient.

For the non-scam call (Table~17), the local score spikes to 0.7051 only in the very first window, which happens to contain a phone-number exchange---a superficial pattern that co-occurs with scam openings in the training distribution. As the conversation moves into unrelated resolution dialogue (apologizing, suggesting a phone directory, saying goodbye), the score decays by two to three orders of magnitude within four to six seconds (windows 4.0--16.0 s) and remains near zero thereafter. This confirms that the detector reacts to local lexical/acoustic surface features within its bounded window rather than propagating a persistent belief state about the call, and that any such false activation is short-lived and is down-weighted by the attention-based aggregator (Eq.~14--15) at the conversation level.

For the scam call (Table~18), scores exceed 0.95 in windows that restate one of a small set of lexical anchors---\textit{suspended}, \textit{verify}, \textit{national security}, \textit{social security number}---even across non-contiguous windows (e.g., 0.0--10.0 s, 6.0--16.0 s, 14.0--24.0 s, 34.0--50.0 s). Notably, the model's sensitivity appears keyed to the literal recurrence of these terms rather than to the underlying scam intent: window 12.0--22.0 s, which paraphrases the same suspension claim as \textit{``This suspension is part of a national-level investigation''} without repeating the anchor phrase verbatim, drops to 0.3761 even though it is flanked by windows scoring above 0.95 on either side. A similar pattern appears later in the call: once the dialogue shifts into negotiation and stalling (callback-number requests, checking with a supervisor, procedural questions about visiting an office), scores collapse to below 0.05 for several consecutive windows (64.0--106.0 s) despite the manipulative intent of the conversation persisting throughout, then rebound sharply as soon as the caller reiterates the identity-verification request (112.0--122.0 s, 0.9922). This intermittency suggests that StreamFraudNet's window-level scores behave more like a keyword-triggered detector operating within a short temporal horizon than a model that tracks cumulative deceptive framing across the call.

These qualitative patterns are consistent with the quantitative findings in Section~4.3 and the paper's limitations: recurrent context within a window is the dominant architectural contributor (Table~3), but the BiLSTM's temporal reach is confined to the current $k$-chunk window and has no mechanism for carrying evidence forward once a window slides out of context. Because no timestamped fraud annotations are available during training (Section~3.1), these local scores are optimized purely to support the conversation-level BCE objective and should therefore be read as latent risk indicators rather than validated localizations of fraudulent intent---the same caveat raised in the paper's Limitations section, and one that these appendix examples make concrete rather than abstract.

\begin{table*}[htbp]
\centering
\caption{Example of a Non-Scam Conversation}
\label{tab:non_scam_example}
\resizebox{\textwidth}{!}{%
\begin{tabular}{ll}
\hline
\textbf{Speaker} & \textbf{Utterance} \\
\hline
Caller & How is this 5551234? \\
Receiver & No, this is 5555678. \\
Caller & Oh, I'm so sorry. I must have dialed the wrong number. \\
Receiver & That's okay. It happens. Who were you trying to reach? \\
Caller & My friend John. Do you know him? \\
Receiver & No, I don't know anyone by that name. \\
Receiver & You might want to try checking the phone book or online directory to get the correct number. \\
Caller & Okay, thanks for your help. \\
Caller & Sorry again for disturbing you. \\
Receiver & No problem. Have a good day. \\
\hline
\end{tabular}%
}
\end{table*}

\begin{table*}[htbp]
\centering
\caption{Example of a Scam Conversation}
\label{tab:scam_example}
\resizebox{\textwidth}{!}{%
\begin{tabular}{ll}
\hline
\textbf{Speaker} & \textbf{Utterance} \\
\hline
Caller & Hello, my name is Officer Johnson from the Social Security Administration. \\ & We've been trying to reach you about a serious issue with your social security number. \\
Receiver & Oh really? What's going on with it? \\
Caller & It's been suspended due to suspicious activity. I need to verify some information to reactivate it. \\
Receiver & Suspended? That sounds serious. Can you tell me more about this suspicious activity? \\
Caller & I'm not at liberty to disclose that information over the phone.\\ 
& But I can assure you it's a matter of national security. \\
Receiver & National security? You're scaring me. What do I need to do to fix this? \\
Caller & I just need you to confirm your social security number and we can move forward with the process. \\
Receiver & Confirm my social security number? Why do you need that? Can't you just look it up in your system? \\
Caller & Our system is down for maintenance. It's a one-time verification process. I assure you. \\
Receiver & Okay, but how do I know you're really from the Social Security Administration? \\ 
& Can you give me a callback number or something? \\
Caller & I'm happy to provide you with a callback number. Let me check with my supervisor real quick. \\ 
& Hold on for just a second. \\
Receiver & Take your time. I'm curious. What's the protocol for dealing with suspended social security numbers? \\ 
& Is there a website or an office I can visit? \\
Caller & No, no, no. This is a highly sensitive matter. You need to deal directly with me or risk further complications. \\
Caller & Now, are you ready to proceed with the verification process? \\
Receiver & Not so fast. You said you were going to give me a callback number. \\
Caller & Right, right. The number is 202-555-1234. You can call back and ask for Officer Johnson. \\
Receiver & Okay, got it. But before we proceed, can you explain what exactly will happen once \\ 
& my social security number is verified? \\
Caller & Once it's verified, we'll reactivate your account and send you a new card in the mail. \\
Receiver & That sounds too good to be true. How much does this service cost? \\
Caller & It's absolutely free. We are doing this as a courtesy to our citizens. \\
Receiver & Free? Really? I'm not buying it. \\
\hline
\end{tabular}%
}
\end{table*}

\begin{table*}[t]
\centering
\caption{Model Scores over Sliding Windows for a Fraudulent Conversation in Table~\ref{tab:non_scam_example}}
\label{tab:scores_table5}
\begin{tabular}{lp{11cm}c}
\hline
\textbf{Window(s)} & \textbf{Transcript} & \textbf{Score} \\
\hline
0.0--10.0  & How is this 5551234? No, this is 5555678. Oh, I'm so sorry. I must have dialed the wrong number. & 0.7051 \\
2.0--12.0  & No, this is 555, 5678. Oh, I'm so sorry. I must have dialed the wrong number. That's okay. It has. & 0.0720 \\
4.0--14.0  & 555-5678. Oh I'm so sorry. I must have dialed the wrong number. That's okay, it happens. Who were you trying & 0.1380 \\
6.0--16.0  & I'm so sorry. I must have dialed the wrong number. That's okay. It happens. Who were you trying to reach? & 0.0059 \\
8.0--18.0  & Sorry. I must have dialed the wrong number. That's okay. It happens. Who were you trying to reach? My friend John. Do you know him? & 0.0045 \\
10.0--20.0 & That's okay, it happens. Who were you trying to reach? My friend John, do you know him? No, I don't know anyone by that name. & 0.0011 \\
12.0--22.0 & Happens. Who were you trying to reach? My friend John. Do you know him? No, I don't know anyone by that name. You might want to try to & 0.0006 \\
14.0--24.0 & To reach. My friend John, do you know him? No, I don't know anyone by that name. You might want to try checking the phone book or online directory & 0.0023 \\
16.0--26.0 & John. Do you know him? No, I don't know anyone by that name. You might want to try checking the phone book or online directory to get the correct number. & 0.2517 \\
18.0--28.0 & No, I don't know anyone by that name. You might want to try checking the phone book or online directory to get the correct number. Okay, thanks for your help. & 0.0109 \\
20.0--30.0 & You might want to try checking the phone book or online directory to get the correct number. Okay, thanks for your help. Sorry again for disturbing you. & 0.1265 \\
22.0--32.0 & Checking the phone book or online directory to get the correct number. Okay, thanks for your help. Sorry again for disturbing you. No problem, have a good day. & 0.0001 \\
24.0--34.0 & To get the correct number. Okay, thanks for your help. Sorry again for disturbing you. No problem. Have a good day. & 0.0000 \\
\hline
\end{tabular}
\end{table*}

\begin{table*}[t]
\centering
\caption{Model Scores over Sliding Windows for a Fraudulent Conversation in Table~\ref{tab:scam_example}}
\label{tab:scores_table6}
\begin{tabular}{lp{11cm}c}
\hline
\textbf{Window (s)} & \textbf{Transcript} & \textbf{Score} \\
\hline
0.0--10.0   & Hello, this is Officer Johnson from the Social Security Administration. We've detected an issue with your SSN. & \textbf{1.0000} \\
4.0--14.0   & We've been trying to reach you regarding suspicious activity linked to your social security number. & 0.6239 \\
6.0--16.0   & Your social security number has been suspended temporarily due to a serious issue. & \textbf{0.9996} \\
12.0--22.0  & This suspension is part of a national-level investigation. We need to verify your identity. & 0.3761 \\
14.0--24.0  & Please confirm some details so we can reactivate your number and prevent legal complications. & \textbf{0.9921} \\
22.0--32.0  & Can you explain the suspicious activity? --- Sorry, that's confidential due to national security concerns. & \textbf{0.9867} \\
26.0--36.0  & While I can't share details, I can assure you this is a critical and urgent matter. & \textbf{0.9546} \\
34.0--44.0  & We need your social security number to proceed with identity verification. & \textbf{1.0000} \\
40.0--50.0  & Once verified, your account will be reactivated and any holds will be removed. & \textbf{1.0000} \\
52.0--62.0  & The system is undergoing maintenance, so manual confirmation is required at this time. & 0.0012 \\
54.0--64.0  & How can I trust this call is legitimate? Can you prove your identity? & \textbf{0.9988} \\
64.0--74.0  & I'd like a callback number or official reference before giving you any personal information. & 0.0515 \\
68.0--78.0  & Let me ask my supervisor. Please hold on for just a moment while I confirm. & 0.0004 \\
76.0--86.0  & In the meantime, what's the standard protocol for dealing with a suspended SSN? & \textbf{0.9996} \\
82.0--92.0  & This case is too sensitive for normal procedures like websites or walk-in visits. You must deal with me directly. & 0.1586 \\
88.0--98.0  & Delays could result in legal consequences. Let's continue. Are you ready to proceed with the verification now? & 0.0040 \\
96.0--106.0 & Wait. You promised a callback number. Where is it? & 0.0009 \\
112.0--122.0 & After verification, your SSN will be cleared and a new card issued via certified mail. & \textbf{0.9922} \\
132.0--142.0 & This service is completely free. We're doing this to protect citizens like you. --- I'm not convinced. & 0.0001 \\
\hline
\end{tabular}
\end{table*}